%% file: main.tex
\documentclass{article}  %
\usepackage{iclr2026_conference,times}

\input{math_commands.tex}

\usepackage{capt-of, eucal}
\usepackage{xspace}
\usepackage{xcolor}
\usepackage{enumitem}
\usepackage{amsmath,amsthm,amssymb,amsfonts,dsfont,pifont,bm,bbm,mathrsfs,mathtools,nicefrac,extarrows,relsize}
\usepackage{algorithm,algpseudocode,listings}
\usepackage{booktabs,multirow,adjustbox,diagbox,threeparttable,tabularray,setspace}
\usepackage{wrapfig}
\usepackage{graphicx}
\usepackage{subcaption}
\usepackage{tabularx}
\usepackage{array}
\usepackage{makecell}
\usepackage{longtable} 
\usepackage{paracol}
\usepackage{float}
\usepackage{fontawesome5}
\usepackage[tableposition=top]{caption}
\usepackage{colortbl}
\usepackage{color}
\usepackage{multicol}
\usepackage{lipsum}

\definecolor{citeblue}{rgb}{0.21,0.49,0.74}
\usepackage[pagebackref=false,breaklinks,colorlinks,citecolor=citeblue,bookmarks=false]{hyperref}
\usepackage{wrapfig}
\usepackage[capitalize]{cleveref}  %

\crefname{section}{Sec.}{Secs.}
\Crefname{section}{Section}{Sections}
\crefname{appendix}{Appendix}{Appendices}
\Crefname{appendix}{Appendix}{Appendices}
\crefname{table}{Table}{Tables}
\Crefname{table}{Table}{Tables}
\crefname{figure}{Fig.}{Figs.}
\Crefname{figure}{Figure}{Figures}
\crefname{equation}{Eq.}{Eqs.}
\Crefname{equation}{Equation}{Equations}
\crefname{theorem}{Thm.}{Thms.}
\Crefname{theorem}{Theorem}{Theorems}
\crefname{lemma}{Lem.}{Lems.}
\Crefname{lemma}{Lemma}{Lemmas}
\crefname{remark}{Rem.}{Rems.}
\Crefname{remark}{Remark}{Remarks}
\crefname{corollary}{Cor.}{Cors.}
\Crefname{corollary}{Corollary}{Corollaries}
\crefname{algorithm}{Alg.}{Algs.}
\Crefname{algorithm}{Algorithm}{Algorithms}
\definecolor{cellred}{RGB}{213, 123, 101}
\definecolor{cellgreen}{RGB}{0, 205, 0}
\definecolor{cellblue}{RGB}{54, 125, 189}
\definecolor{codegreen}{rgb}{0,0.6,0}
\definecolor{codegray}{rgb}{0.5,0.5,0.5}
\definecolor{codepurple}{rgb}{0.58,0,0.82}
\definecolor{backcolour}{rgb}{1.0,1.0,1.0}
\lstdefinestyle{mystyle}{
    backgroundcolor=\color{backcolour},
    commentstyle=\color{codegreen},
    keywordstyle=\color{magenta},
    numberstyle=\tiny\color{codegray},
    stringstyle=\color{codepurple},
    basicstyle=\ttfamily\scriptsize,
    breakatwhitespace=false,
    breaklines=true,
    captionpos=b,
    keepspaces=true,
    numbers=left,
    numbersep=5pt,
    showspaces=false,
    showstringspaces=false,
    showtabs=false,
    tabsize=2
}
\usepackage[most,skins,theorems]{tcolorbox}
\tcbset{
  aibox/.style={
    width=\linewidth,
    top=8pt,
    bottom=4pt,
    colback=blue!6!white,
    colframe=black,
    colbacktitle=black,
    enhanced,
    center,
    attach boxed title to top left={yshift=-0.1in,xshift=0.15in},
    boxed title style={boxrule=0pt,colframe=white,},
  }
}
\newtcolorbox{AIbox}[2][]{aibox,title=#2,#1}
\usepackage{amsmath} %
\usepackage{amssymb} %
\usepackage{multirow}
\usepackage{comment}
\usepackage{rotating}  %

\newcolumntype{C}[1]{>{\centering\arraybackslash}p{#1}}
\newcolumntype{L}[1]{>{\arraybackslash}p{#1}}

\definecolor{demphcolor}{gray}{.2}

\definecolor{demphcolorinline}{gray}{.3}

\definecolor{demphcolor1}{gray}{.6}

\newcommand{\tocite}[1]{{\color{red} [TO CITE]}}

\title{
    \raisebox{-0.2\height}{\includegraphics[width=1.0cm,height=1.0cm,keepaspectratio]{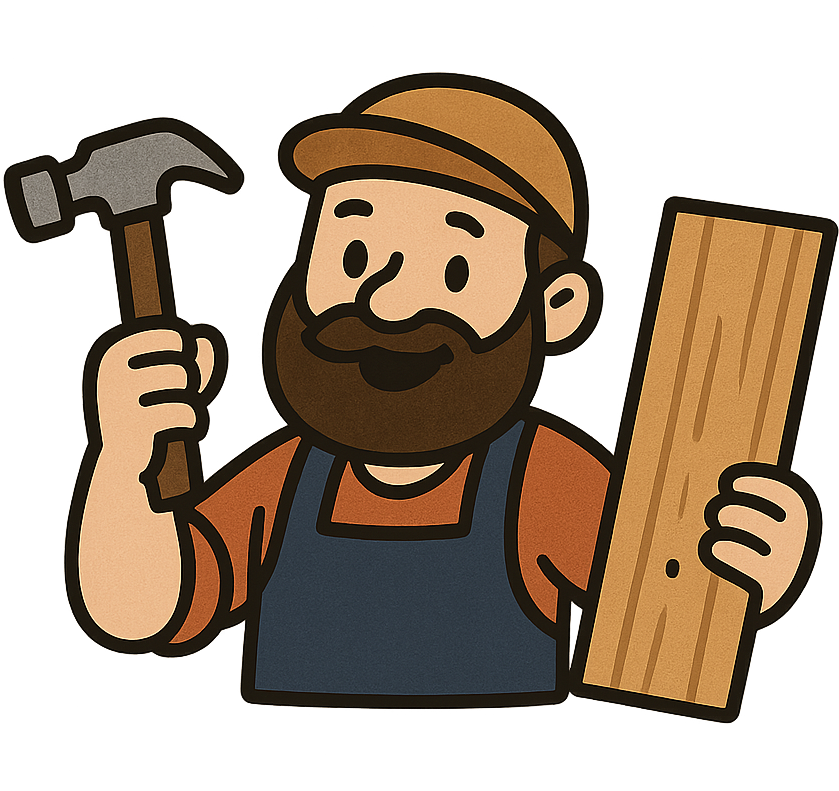}}
    Cooper: Co-Optimizing Policy and Reward Models in Reinforcement Learning for Large Language Models
}

\author{%
  \textbf{Haitao Hong}$^{1,*}$,
  ~~
  \textbf{Yuchen Yan}$^{1,*}$,
  ~~
  \textbf{Xingyu Wu}$^{1}$,
  ~~
    \textbf{Guiyang Hou}$^{1}$,
  ~~ 
    \textbf{Wenqi Zhang}$^{1}$ 
  \\
    \textbf{Weiming Lu}$^{1}$ 
  ~~
    \textbf{Yongliang Shen}$^{1,\dagger}$
  ~~
    \textbf{Jun Xiao}$^{1}$
  \\
  $^1$Zhejiang University\\
  \texttt{\{haitaohong, yanyuchen, syl\}@zju.edu.cn} \\
  \begin{tabular}{@{}ll@{}}
  \vspace{0.1cm} \\
    \faGithub\ GitHub: & \href{https://github.com/zju-real/cooper}{\texttt{\textcolor{cyan}{https://github.com/zju-real/cooper}}} \\
    \faGlobe\ Project: & \href{https://zju-real.github.io/cooper}{\texttt{\textcolor{cyan}{https://zju-real.github.io/cooper}}}
  \end{tabular}
}

\iclrfinalcopy %

\begin{document}

\maketitle

\renewcommand{\thefootnote}{\fnsymbol{footnote}}
\footnotetext[1]{~The first two authors have equal contributions.}
\footnotetext[2]{~Corresponding author.}
\renewcommand{\thefootnote}{\arabic{footnote}}

\input{sections/0.abs}

\input{sections/1.intro}

\input{sections/1-2.related}

\input{sections/2.method2}

\input{sections/3.experiment}

\input{sections/4.results}

\input{sections/5.conclusion}

\input{sections/9.bib}

\appendix
\input{sections/99.appendix}

\end{document}

%% file: math_commands.tex
\usepackage{amsmath,amsfonts,bm}

\def\eqref#1{equation~\ref{#1}}

\def\1{\bm{1}}

\DeclareMathAlphabet{\mathsfit}{\encodingdefault}{\sfdefault}{m}{sl}
\SetMathAlphabet{\mathsfit}{bold}{\encodingdefault}{\sfdefault}{bx}{n}

%% file: sections/0.abs.tex
\begin{abstract}

Large language models (LLMs) have demonstrated remarkable performance in reasoning tasks, where reinforcement learning (RL) serves as a key algorithm for enhancing their reasoning capabilities. Currently, there are two mainstream reward paradigms: model-based rewards and rule-based rewards. However, both approaches suffer from limitations: rule-based rewards lack robustness, while model-based rewards are vulnerable to reward hacking.
To address these issues, we propose \textbf{Cooper} (\underline{Co-o}ptimizing \underline{P}olicy Mod\underline{e}l and \underline{R}eward Model), a RL framework that jointly optimizes both the policy model and the reward model. Cooper leverages the high precision of rule-based rewards when identifying correct responses, and dynamically constructs and selects positive-negative sample pairs for continued training the reward model. This design enhances robustness and mitigates the risk of reward hacking.
To further support Cooper, we introduce a hybrid annotation strategy that efficiently and accurately generates training data for the reward model. We also propose a reference-based reward modeling paradigm, where the reward model takes a reference answer as input. Based on this design, we train a reward model named VerifyRM, which achieves higher accuracy on VerifyBench compared to other models of the same size. We conduct reinforcement learning using both VerifyRM and Cooper. Our experiments show that Cooper not only alleviates reward hacking but also improves end-to-end RL performance, for instance, achieving a 0.54\% gain in average accuracy on Qwen2.5-1.5B-Instruct.
Our findings demonstrate that dynamically updating reward model is an effective way to combat reward hacking, providing a reference for better integrating reward models into RL.

\end{abstract}

%% file: sections/1.intro.tex
\section{Introduction}

\begin{quote}
    \textit{``He who teaches, who learns.''} 
    \begin{flushright}
        --- Confucius
    \end{flushright}
\end{quote}

Large language models (LLMs) have demonstrated remarkable capabilities in reasoning~\citep{kumar2025llm,openai2024gpt4ocard}, particularly in mathematical reasoning~\citep{lewkowycz2022solving,shao2024deepseekmath,ying2024internlm,yang2024qwen2}, code reasoning~\citep{roziere2023code,zhu2024deepseek,hui2024qwen2}, and commonsense reasoning~\citep{wang2024candle}, often achieving or even surpassing human-level performance. Recent advances demonstrate that reinforcement learning (RL) has become a pivotal technique for enhancing these reasoning capabilities \citep{xu2025towards}. By generating multiple solution trajectories and optimizing the model to align with high-quality responses, RL enables LLMs to achieve performance that often matches or exceeds human expertise \citep{havrilla2024teaching}.

\begin{figure}[t]
    \centering
    \includegraphics[width=0.7\linewidth]{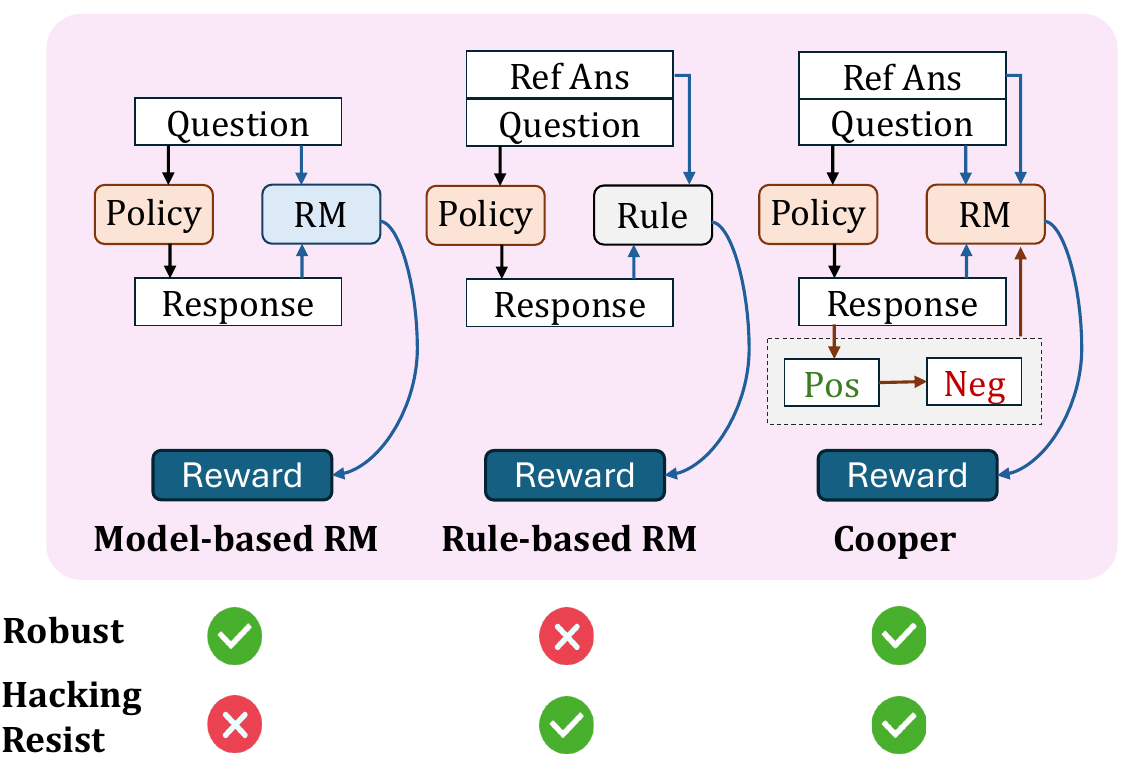}
    \caption{Model-based rewards are generally robust to variations in model outputs, but they are susceptible to being hacked by the policy model. In contrast, rule-based rewards are less prone to hacking but often lack robustness. We introduce \textbf{Cooper}, a reinforcement learning framework that achieves both high robustness and resistance to reward hacking. In this figure, the black arrows indicate the rollout process, the blue arrows represent the reward assignment process, and the brown arrows denote the update process for the reward model.}
    \label{fig:enter-label}
\end{figure}

In RL algorithms, one critical factor affecting performance is the design of the reward function, as it determines the quality of evaluation for output sequences. In the early stages of RL for LLMs, human preference data was typically used to train a reward model that assigns rewards based on the input question and model generations~\citep{ouyang2022training,bai2022training}. This paradigm, known as reinforcement learning from human feedback (RLHF), has been widely adopted~\citep{havrilla2024teaching,DBLP:journals/corr/abs-2503-04548}. Since the introduction of reasoning models like OpenAI's o1~\citep{jaech2024openai}, DeepSeek-R1~\citep{guo2025deepseek} and Kimi-k1.5~\citep{kimiteam2025kimik15scalingreinforcement}, RL for LLMs has shifted focus towards verifiable tasks, where rule-based reward functions are commonly employed to assign scores, thereby improving the reliability of the scoring system.

However, both model-based and rule-based reward functions have inherent limitations. Model-based rewards, which rely on dynamic calculations based on the model's parameters, are prone to reward hacking~\citep{gao2023scaling}. Specifically, when a fixed reward model is used, the model may exploit output patterns that deceive the reward function, thereby obtaining high scores regardless of the correctness of the output. This phenomenon can lead to catastrophic failures in the later stages of training. On the other hand, rule-based reward functions often rely on manually crafted rules to parse and verify the model's output~\citep{gandenberger2024mathverify,eval-harness}. This method lacks robustness and is susceptible to misjudgment, which constrains the further optimization of the model~\citep{DBLP:conf/nips/ChristianoLBMLA17}.

In our preliminary experiments, we observed that rule-based reward functions exhibit high precision in identifying correct samples and high recall in detecting incorrect ones. That is, samples judged as correct by rule-based reward functions are usually indeed correct, whereas those judged as incorrect may still be correct in reality. This phenomenon arises because rule-based functions are typically handcrafted and lack robustness in answer extraction. As a result, they often fail to accurately extract and match answers, especially when facing diverse output formats generated by different models. 

In this paper, we propose \textbf{Cooper}, a novel reinforcement learning framework that enables synchronized co-optimization of the policy model and the reward model. Cooper introduces a two-stage training pipeline: \textbf{(1) Policy model optimization}: Policy model optimization follows the GRPO paradigm, involving sampling and scoring responses with a reference-aware reward model, and performing policy updates based on within-group normalized advantages and KL regularization; \textbf{(2) Reward model optimization}: Reward model optimization continuously refines the reward model via contrastive learning, using positive samples identified by high-precision rule-based signals and negative samples generated by transforming correct responses into incorrect ones with an assistant LLM.

To support cooper framework, we first focus on training an accurate and robust reference-based reward model. We construct a large-scale dataset using responses generated from diverse LLMs across multiple high-quality math reasoning datasets. A hybrid annotation strategy is applied using both rule-based verifier tools (e.g., Math-Verify) and LLM-based verifiers, allowing for automatic correctness labeling at scale. Using this labeled data, we train a reward model \textbf{VerifyRM} that scores responses, with the query and a reference answer as its input.

We then integrate this reward model into the Cooper pipeline and validate its effectiveness through comprehensive experiments. Our results demonstrate that applying this co-optimization framework significantly improves the policy model’s reasoning ability. Our experiments show that models trained with Cooper outperform both rule-based and fixed-reward-model RL baselines across several challenging math reasoning benchmarks.

The main contributions are summarized as follows:
\begin{itemize}
\item We introduce a reward modeling dataset, which is labeled using a hybrid annotation strategy that combines rule-based verification and LLM-as-a-judge verification, enabling efficient and reliable correctness supervision. The reward model trained on our constructed dataset achieves an accuracy of 89.42\% on VerifyBench, surpassing existing reward models of the same scale. 
\item Based on the high precision of rule-based rewards in identifying correct answers, we propose \textbf{Cooper}, a reinforcement learning framework that co-optimizes the policy model and the reward model simultaneously. This framework mitigates the problem of reward hacking commonly observed in reward-model-based RL and enhances overall training performance. 
\item Our work demonstrates that dynamically adjusting the parameters of the reward model during the RL training process can effectively mitigate the phenomenon of reward hacking, providing valuable insights for the research community on how to better utilize reward models in reinforcement learning.
\end{itemize}

%% file: sections/1-2.related.tex
\section{Related Works}
\paragraph{Reinforcement Learning for Large Language Models. } 

Reinforcement Learning (RL) has emerged as a foundational method for aligning large language models~\citep{DBLP:conf/nips/ChristianoLBMLA17,ziegler2020finetuninglanguagemodelshuman,DBLP:journals/corr/abs-2312-14925}. Early work such as InstructGPT~\citep{ouyang2022training} demonstrated that fine-tuning LLMs with a reward model trained on human preference data can significantly improve response helpfulness and alignment.
However, RLHF is often computationally expensive, costly, and reliant on large-scale human annotations.
To mitigate these issues, recent methods have proposed simplifying the RLHF process.
Direct Preference Optimization (DPO)~\citep{DBLP:conf/nips/RafailovSMMEF23} reformulates the RLHF objective as a contrastive loss over preference pairs, eliminating the need for reward model training and sampling-based updates. Reinforcement Learning with AI Feedback (RLAIF)~\citep{bai2022training} proposes replacing human preference data with AI-generated feedback guided by predefined principles instead of humans, significantly lowering the annotation cost and improving scalability.

\paragraph{Reward Models for Reinforcement Learning. }

Reward Models (RMs) has been widely used in  reasoning tasks~\citep{zhong2025comprehensivesurveyrewardmodels} for reinforcement learning~\citep{ouyang2022training,lambert2024rewardbenchevaluatingrewardmodels,guo2025rewardreasoningmodel} and verification-guided inference~\citep{guo2025rewardreasoningmodel,DBLP:conf/acl/LiLZFCLC23,zhang2025r1rewardtrainingmultimodalreward,DBLP:conf/naacl/YuGW24}.
According to reward modeling mechanisms, Existing RMs broadly fall into three types: (1) discriminative reward models~\citep{DBLP:journals/corr/abs-2403-17297,DBLP:conf/acl/ZangD0CLDWMDZCL25}, typically implemented as classifiers over response sequences, assigning binary or fine-grained preference scores; (2) generative reward models \citep{DBLP:journals/corr/abs-2504-02495,DBLP:journals/corr/abs-2501-17195,hong2025thinkrmenablinglonghorizonreasoning}, which generate textual feedback or critiques before producing a scalar reward; and (3) implicit reward models~\citep{DBLP:journals/corr/abs-2411-15124}, often optimized via DPO~\citep{DBLP:conf/nips/RafailovSMMEF23}, where model likelihoods are interpreted as reward signals. Orthogonally, reward models can be categorized into outcome reward models (ORMs)~\citep{DBLP:journals/corr/abs-2410-18451,cobbe2021gsm8k}, which assign scalar feedback to final outputs, and process reward models (PRMs)~\citep{DBLP:conf/iclr/SetlurNFGEAABK25}, which evaluate intermediate reasoning steps to provide denser and more interpretable supervision. Our VerifyLM belongs to the discriminative RM and ORM. 
\paragraph{Reinforcement Learning with Verifiable Rewards. }As an alternative to learned reward models, Reinforcement Learning with Verifiable Rewards (RLVR)~\citep{guo2025deepseek,DBLP:journals/corr/abs-2411-15124,DBLP:journals/corr/abs-2504-13837} leverages rule-based verification functions—such as exact answer matching or logical consistency checks—to generate reward signals automatically. Notably,  DeepSeek-R1~\citep{guo2025deepseek} achieves strong reasoning performance through a multi-stage pipeline combining supervised pretraining with Group Relative Policy Optimization (GRPO)~\citep{shao2024deepseekmath}. Cooper draws inspiration from the recent advancements in enhancing reasoning through RLVR, screening out the correct responses of the policy model via symbolic verification, and thereby constructing preference data to update the reward model.

%% file: sections/2.method2.tex
\section{Methods}

Our methods consist of two main components. The first part proposes a pipeline for constructing a reference-based reward model \textbf{VerifyRM}, which includes data collection and annotation strategies, as well as the training procedure for the reward model. The second part presents \textbf{Cooper}, a reinforcement learning algorithm that \underline{co-o}ptimizes both the \underline{p}olicy mod\underline{e}l and the \underline{r}eward model. In this framework, the RM trained in the first stage guides the policy model's updates within Cooper while being updated itself concurrently.

\subsection{Training Recipe of VerifyRM}
\label{sec:rm_train_method}
Most existing reward models score the input-output pairs of large language models (LLMs) directly~\citep{zhong2025comprehensivesurveyrewardmodels}. However, in reasoning tasks, there typically exists a reference answer. \citet{yan2025verifybenchbenchmarkingreferencebasedreward} have demonstrated the importance of reference answers for model-based verification. Therefore, we propose a method for constructing reference-based reward models to improve the accuracy of reward models in reasoning tasks.  
\subsubsection{Data Preparation}
To train VerifyLM, the required data format consists of a reasoning problem, its corresponding reference answer, a model-generated completion, and a label indicating whether the completion is correct. 

\begin{figure*}[t!]
    \centering
    \includegraphics[width=\linewidth]{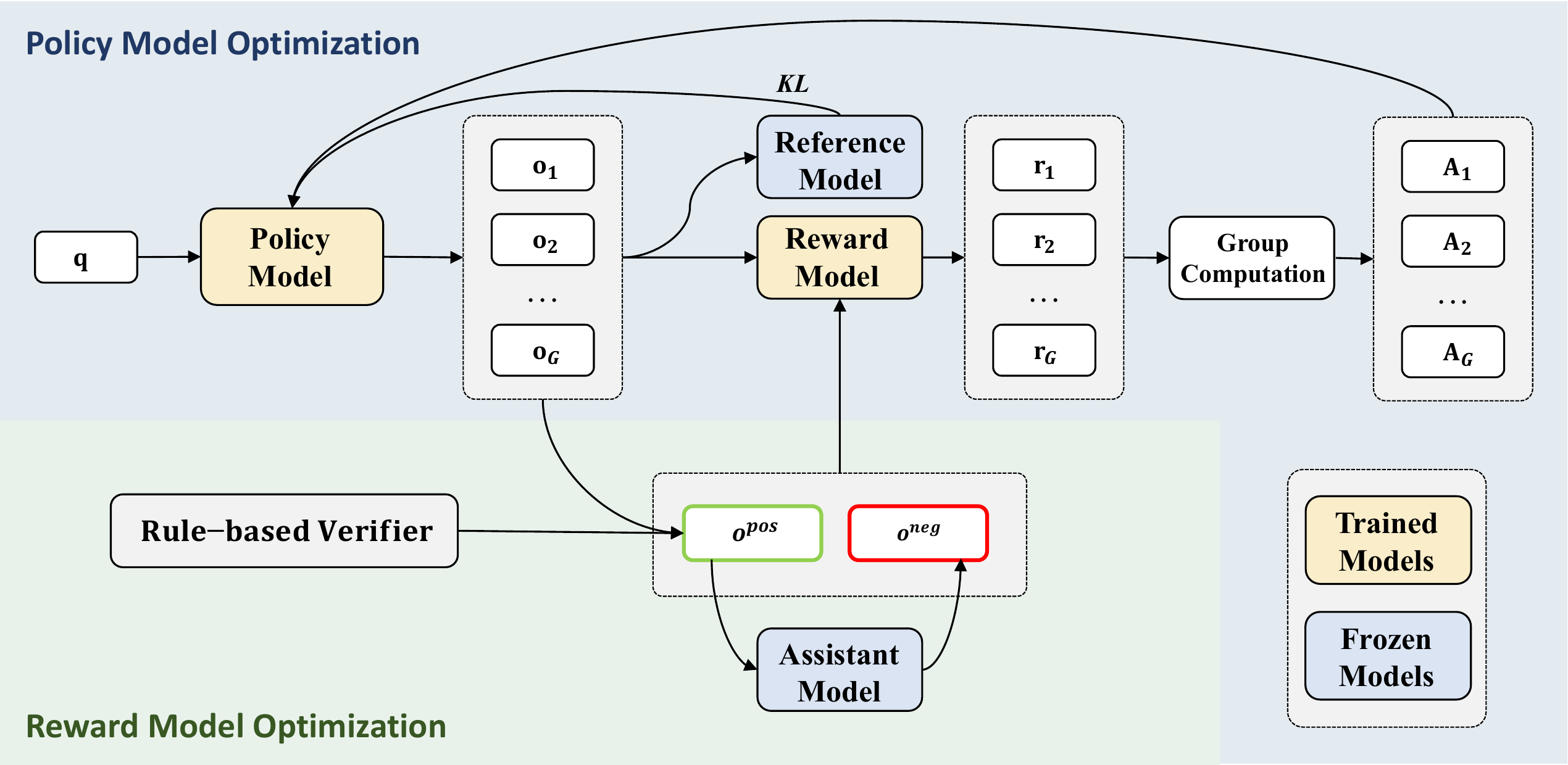}
    \caption{An overview of the Cooper training framework. Each training step in Cooper consists of two stages: policy model optimization (blue area) and reward model optimization (green area).}
    \label{fig:cooper_framework}
\end{figure*}

\paragraph{Problem-reference-completion triples collection.} 
We collected 7 commonly used mathematical reasoning datasets, each containing math problems and their corresponding reference answers. Using 11 mainstream LLMs, we generated completions for these math problems, with each model providing one completion per problem. During sampling, we set the temperature to 0.7 and top\_p to 0.95. In total, we collected 65K problem-reference-completion triples. Details of the datasets, LLMs, and their licenses are provided in the Appendix~\ref{sec:Data Source} and~\ref{sec:LLM Usage}. 

\paragraph{Hybrid labeling for correctness.} 
In prior works, researchers have relied heavily on manual annotation to determine the correctness of model completions~\citep{xVerify}. We observe that current LLMs have already demonstrated strong capabilities in evaluating the correctness of completions against reference answers (e.g., Qwen3-4B achieves 94.17\% accuracy on VerifyBench)~\citep{yang2025qwen3technicalreport}. Motivated by this, we propose an automated hybrid labeling approach that combines a rule-based verifier and an LLM-as-a-judge. Specifically, we use Math-verify~\citep{gandenberger2024mathverify} as the rule-based verifier and Qwen3-4B (in non-thinking mode)~\citep{yang2025qwen3technicalreport} as the LLM-as-a-judge. We only retain samples for which both methods agree on the correctness label, resulting in a dataset of 58.7K examples for training VerifyRM. Detailed statistics are provided in the Appendix~\ref{sec:Details of Hybrid Annotation}.

\subsubsection{Reward Model Training}

In this paper, following the approach of~\citet{DBLP:journals/corr/abs-2403-17297}, the reward model is formulated as a text classifier. However, we incorporate the reference answer into the input of the reward model. The specific input template is provided in the appendix~\ref{sec:Prompt Template for VerifyRM}. Inspired by ~\citet{DBLP:conf/acl/ZangD0CLDWMDZCL25}, we initialize our model from an already aligned LLM, replacing its original language modeling head with a newly initialized score head. The model is then trained using binary cross-entropy loss. The objective function can be formally written as:  
\begin{equation}
\mathcal{L}(\theta) = \mathbb{E}_{[\{q,r,c,y\} \sim D]} \text{BCE}(\sigma({M_{\theta}(q,r,c)}), y)
\end{equation}
\begin{equation}
\text{BCE}(\hat{y}, y) = -y* \log \hat{y} - (1 - y) \log (1 - \hat{y})
\end{equation}
where \( q \) denotes the question, \( r \) denotes the reference answer, \( c \) denotes the model's completion, \( y \) indicates the correctness label, and \( D \) represents the training dataset. The sigmoid function \( \sigma \) is used to map the logits output by the model \( M_\theta \) into the range \([0, 1]\), which is then used to compute the binary cross-entropy loss.

\subsection{Reinforcement Learning with Cooper}

\begin{algorithm*}[t!]
\small
\caption{\textbf{Co-Optimizing Policy and Reward Models (Cooper)}}
\label{alg:cooper}
\begin{algorithmic}
\State \textbf{Input:} initial policy model $\pi_{\theta_{\text{init}}}$, reward models $R_{\varphi_\text{init}}$, training data $\mathcal{D}$; hyperparameters $\varepsilon, \beta$ 
\State\textbf{Output: $\pi_{\theta}^*$ and $R_{\varphi}^*$ } 

\State policy model $\pi_\theta \gets \pi_{\theta_{\text{init}}}$
reward model $R_\varphi \gets R_{\varphi_{\text{init}}}$

\For{iteration $i = 1$ to $I$ }
    \State reference model $\pi_{\text{ref}} \gets \pi_{\theta}$
    \State Sample a batch $\mathcal{D}_b \subset \mathcal{D}$
    \For{each question and answer $(q,a) \in \mathcal{D}_b$}
        \State Generate $G$ rollouts: $\{o_1, \ldots, o_G\}$ using $\pi_{\theta}$ \hfill 
        \State Compute rewards by running $r_{\varphi}$ on each output $o_i$:  $r_j = \sigma(R_{\varphi}(q, o_j, a))$
        \State Compute $\hat{A}_{i,t}$ for the $t$-th token of $o_i$ through group relative advantage estimation:
        \[ \hat{A}_{i,t}=\frac{r_{i}-\operatorname{mean}\left(\left\{r_{1}, r_{2}, \cdots, r_{G}\right\}\right)}{\operatorname{std}\left(\left\{r_{1}, r_{2}, \cdots, r_{G}\right\}\right)} .\]
        \State  Select a positive response \( o_{\text{pos}}  \in \{o_1, \ldots, o_G\} \) using rule-based reward function: \hfill 
        \[o_{\text{pos}} = o \sim \{o_i \mid \text{Rule}(a, o_i) = 1 \}\]
        \State Generate a negative response \( o_{\text{neg}} \) from an assistant LLM: \hfill
        \[o_{\text{neg}} = M(p, o_{\text{pos}})\]
        \EndFor

    \State Update the policy model $\pi_\theta$ by maximizing the GRPO objective:              \hfill \textbf{$\triangleright$ Stage 1: Policy model optimization}
\[
\begin{aligned}
\mathcal{J}_{\text{GRPO}}(\theta) 
&= \mathbb{E}\left[
    q \sim P(Q),
    \left\{o_i\right\}_{i=1}^G \sim \pi_{\theta_{\text{old}}}(O \mid q)
\right] \frac{1}{G} \sum_{i=1}^G \frac{1}{|o_i|} \sum_{t=1}^{|o_i|}\\
& \bigg\{
\min \Bigg[
\frac{
    \pi_{\theta}(o_{i, t}\mid q, o_{i,<t})
}{
    \pi_{\theta_{\text{old}}}(o_{i, t} \mid q, o_{i,<t})
} \hat{A}_{i,t},
\operatorname{clip}\left(
\frac{
    \pi_{\theta}(o_{i, t} \mid q, o_{i,<t})
}{
    \pi_{\theta_{\text{old}}}(o_{i, t} \mid q, o_{i,<t})
}, \varepsilon
\right) \hat{A}_{i,t}
\Bigg]
- \beta\, \mathbb{D}_{\mathrm{KL}}\left[\pi_\theta \parallel \pi_{\text{ref}}\right]
\bigg\}
\end{aligned}
\]

    \State Update the reward model $ R_\varphi$ by minimizing the loss: \hfill \textbf{$\triangleright$ Stage 2: Reward model optimization}
    \[\begin{split}
    \mathcal{L}_{\mathrm{RM}} &= -\mathbb{E}_{[\{q, a, o_{\text{pos}}, o_{\text{neg}} \} \sim \mathcal{D}_b]} log \sigma(R_\varphi(q,a,o_{\text{pos}}) - R_\varphi(q,a,o_{\text{neg}}))
    \end{split}\]
\EndFor
\State \textbf{return} $\pi_{\theta}^*$ and $R_{\varphi}^*$
\end{algorithmic}
\end{algorithm*}

We propose \textbf{Cooper}, a reinforcement learning framework that co-optimizes policy and reward models. Cooper enables simultaneous tuning of the policy model and reward model in a single training step. We present Cooper in Figure~\ref{fig:cooper_framework} and Algorithm ~\ref{alg:cooper}.

\paragraph{Stage 1: Policy model optimization.} 

Following the GRPO~\citep{shao2024deepseekmath} paradigm, our policy model optimization proceeds as follows. For each training sample \(q\), we sample a set of responses \(\{o_1, o_2, \ldots, o_n\}\) using the policy\(\pi_\theta\). The reward model then evaluates each rollout, producing scores \(\{r_1, r_2, \ldots, r_n\} \). We normalize these rewards across the group to compute advantage estimates \(\{A_1, A_2, \ldots, A_n\}\), which are subsequently used to update the policy via policy gradient. To regularize exploration and ensure training stability, a KL divergence penalty is incorporated during reinforcement learning.  

However, unlike previous RL methods based on reward models, we incorporate a reference answer, denoted as \(a\), into the scoring process of the reward model. Consequently, the reward \(r\) can be computed as: 
\begin{equation}
    r_i = R_\varphi(q, a, o_i)
\end{equation}
The computation of the remaining variables and the optimization of the policy model follow the same methodology as proposed in \citet{shao2024deepseekmath}. 
\paragraph{Stage 2: Reward model optimization.} 
Cooper introduces a new step into the existing GRPO pipeline: the optimization of the reward model. This is designed to ensure that the RM’s parameters are continuously updated during the RL process, thereby reducing the risk of the policy model exploiting specific vulnerabilities in the RM (i.e., reward hacking) and ultimately maintaining the stability of training.  

Specifically, following the approach of ~\citet{NEURIPS2020_1f89885d}, we optimize the reward model using contrastive learning. Given a question \( q \), a reference answer \( a \), and a pair of candidate responses \( o_{\text{pos}} \) and \( o_{\text{neg}} \) to \( q \), where \( o_{\text{pos}} \) is a correct response and \( o_{\text{neg}} \) is incorrect, the objective is to maximize the score difference assigned by the RM between \( o_{\text{pos}} \) and \( o_{\text{neg}} \). The optimization could be represented as:
\begin{equation}
\label{eq:reward_loss}
\begin{split}
\mathcal{L}_{\mathrm{RM}} &= -\mathbb{E}_{[\{q, a, o_{\text{pos}}, o_{\text{neg}} \} \sim D]} log \sigma(R_\varphi(q,a,o_{\text{pos}}) - R_\varphi(q,a,o_{\text{neg}}))
\end{split}
\end{equation}
To obtain a set of positive and negative samples \(o_{\text{pos}}, o_{\text{neg}}\) for a given question \(q\) and its reference answer \(a\), we perform the following operations respectively: 

\paragraph{Positive sample selection.} Based on our preliminary observations of rule-based rewards, we found that such rules tend to exhibit high precision but low recall in identifying correct responses. In other words, responses classified as correct by the rule are highly likely to be truly correct. Therefore, for a single rollout that yields a set of responses \(\{o_1, o_2, ..., o_n\}\), we randomly select one response that is judged as correct by the rule and treat it as a positive sample:
\begin{equation}
    o_{\text{pos}} = o \sim \{o_i \mid \text{Rule}(a, o_i) = 1 \}
\end{equation}
\paragraph{Negative sample generation.} We propose a simple method for generating negative samples. Specifically, we utilize an assistant LLM \(M\) to transform a correct reasoning process into one that ultimately yields an incorrect answer, guided by a carefully designed prompt \(p\) (shown in Appendix~\ref{sec:Prompt Template for generating negative response}). To ensure the generated response is indeed incorrect, we incorporate a verification mechanism. Leveraging the high precision of a rule-based reward system, we pass the generated reasoning process through the rule-reward to verify its correctness. If it is not identified as incorrect by the rule-reward, the process is repeated until a valid negative sample is obtained:
\begin{equation}
    o_{\text{neg}} = M(p, o_{\text{pos}})
\end{equation}
We would like to mention that if no valid pair is constructed during our data construction process, we add a loss mask to skip the optimization for that sample.

%% file: sections/3.experiment.tex
\section{Experiments}
\subsection{Preliminary Experiment}

To validate our hypothesis that rule-based verifiers exhibit high precision despite low recall, we analyzed the verification patterns of Math-Verify~\citep{gandenberger2024mathverify} and Qwen3-4B~\citep{yang2025qwen3technicalreport} on VerifyBench~\citep{yan2025verifybenchbenchmarkingreferencebasedreward}. Table~\ref{tab:confusion_matrix} reveals clear asymmetry: Math-Verify achieves 96\% precision (345/360) when identifying correct responses but only 63\% recall (345/549), while Qwen3-4B shows balanced performance with 90\% precision and 99\% recall. This reflects Math-Verify's conservative parsing, which only accepts responses with clearly extractable answers in expected formats, rejecting many correct solutions with non-standard presentations.

\begin{wraptable}[9]{r}{0.53\textwidth}
  \vspace{-2ex} 
  \centering
  \small
  \setlength{\tabcolsep}{1pt}
\begin{tabular}{lrrrr}
    \toprule
    \multicolumn{1}{l}{\textbf{VerifyBench}} & \multicolumn{2}{c}{\textbf{Math-Verify}} & \multicolumn{2}{c}{\textbf{Qwen3-4B}} \\
\cmidrule{2-3} \cmidrule{4-5} \textbf{(Math Reasoning)} & \multicolumn{1}{l}{Pred = 1} & \multicolumn{1}{l}{Pred = 0} & \multicolumn{1}{l}{Pred = 1} & \multicolumn{1}{l}{Pred = 0} \\
    \midrule
    Label = 1 & 345   & 204   & 543   & 6 \\
    Label = 0 & 15    & 534   & 58    & 491 \\
    \bottomrule
    \end{tabular}%
  \caption{Confusion matrices for rule-based (Math-Verify) and model-based (Qwen3-4B) verifiers on VerifyBench.}
  \label{tab:confusion_matrix}%
\end{wraptable}%

This finding directly motivates Cooper's design. The near-perfect precision of rule-based verification when it succeeds provides highly reliable positive signals for training. By using Math-Verify to select positive examples for reward model updates, we leverage its precision while avoiding its recall limitations. Meanwhile, the reward model handles the broader distribution of responses during policy optimization. This complementary approach, combining rule-based precision for reward updates with model-based flexibility for policy scoring, forms the foundation of our co-optimization framework.

\subsection{Experiments for VerifyRM}

\begin{wraptable}[19]{rt}{0.5\textwidth}
  \vspace{-1em} 
  \centering
  \small
    \setlength{\tabcolsep}{0.01pt}
    \resizebox{1.0\linewidth}{!}{
    \begin{tabular}{ll}
    \toprule
    \textbf{Method} & \textbf{VerifyBench-Math} \\
    \midrule
    \rowcolor[rgb]{ .91,  .91,  .91} \multicolumn{2}{l}{\textit{Rule-based function}} \\
    \midrule
    Math-Verify & \multicolumn{1}{r}{79.93} \\
    \midrule
    \rowcolor[rgb]{ .91,  .91,  .91} \multicolumn{2}{l}{\textit{Vanilla reward model w/o reference}} \\
    \midrule
    FsfairX-LLaMA3-RM-v0.1 & \multicolumn{1}{r}{49.53} \\
    Skywork-Reward-V2-Llama-3.2-1B & \multicolumn{1}{r}{47.23} \\
    Skywork-Reward-V2-Llama-3.2-3B& \multicolumn{1}{r}{52.63} \\
    Skywork-Reward-V2-Llama-3.1-8B & \multicolumn{1}{r}{52.06} \\
    Llama-3.1-Tulu-3-8B-RM & \multicolumn{1}{r}{51.56} \\
    \midrule
    \rowcolor[rgb]{ .91,  .91,  .91} \multicolumn{2}{l}{\textit{Reference-based verifier}} \\
    \midrule
    xVerify-0.5B-I & \multicolumn{1}{r}{70.68} \\
    xVerify-3B-Ia & \multicolumn{1}{r}{82.23} \\
    xVerify-8B-I & \multicolumn{1}{r}{84.38} \\
    xVerify-9B-C & \multicolumn{1}{r}{84.23} \\
    \textbf{VerifyRM-1.5B (ours)} & \multicolumn{1}{r}{\textbf{89.42}} \\
    \bottomrule
    \end{tabular}%
    }
  \vspace{1em}
  \caption{Reward model accuracy on VerifyBench.}
  \label{tab:rm_results}%
\end{wraptable}%

We trained VerifyRM following the methodology in Section~\ref{sec:rm_train_method}, using Qwen2.5-Math-1.5B-Instruct~\citep{yang2024qwen2} as the base model. Training was conducted for 3 epochs with a learning rate of 2e-5 and batch size of 128. To ensure fair evaluation on VerifyBench, we excluded all overlapping queries from our training data.

Table~\ref{tab:rm_results} compares VerifyRM against three categories of baselines: rule-based functions, vanilla reward models, and reference-based verifiers. The results demonstrate clear performance stratification. Vanilla reward models without reference answers perform poorly (47-52\% accuracy), confirming that standard preference-based rewards lack the precision needed for mathematical verification. Rule-based Math-Verify achieves 79.93\%, validating its utility but also highlighting its brittleness. Among model-based verifiers, performance scales with model size, yet our VerifyRM-1.5B achieves the highest accuracy at 89.42\%, outperforming even the 9B parameter xVerify model. This superior performance with fewer parameters validates two key design choices: incorporating reference answers provides crucial context for verification, and our hybrid annotation strategy creates higher-quality training data than existing approaches. The strong performance of VerifyRM establishes the reliable reward signal necessary for Cooper's co-optimization framework.

\subsection{Experiments for Cooper}

\paragraph{Setup.}
We implemented the Cooper algorithm based on the veRL~\citep{sheng2024hybridflow} framework. The experiments were conducted on the DeepMath~\citep{he2025deepmath103klargescalechallengingdecontaminated} dataset. Due to resource constraints, we randomly sampled 10K examples from the original dataset for training. All experiments used Qwen2.5-1.5B-Instruct~\citep{qwen2025qwen25technicalreport} and Llama-3.2-1B-Instruct~\citep{grattafiori2024llama3herdmodels} as the initial model. To avoid introducing additional knowledge, the assistant model in Cooper was also instantiated with the same model. In the GRPO algorithm, we set the global batch size to 512, the maximum prompt length to 1024, and the maximum response length to 3072. The learning rate was set to 1e-6, and the KL penalty coefficient was set to 0.001. For each prompt, we generated 16 rollouts during RL training. The models are trained with 10 epochs.

\paragraph{Evaluation.}

We evaluate the model on five mathematical reasoning benchmarks: GSM8K~\citep{cobbe2021gsm8k},  SVAMP~\citep{patel-etal-2021-nlp}, MATH500~\citep{lightman2023lets}, OlympiadBench-EN (OB-EN)~\citep{he2024olympiadbench}, and Math Odyssey~\citep{fang2024mathodyssey}. Among them, GSM8K, MATH500, and SVAMP represent elementary-level mathematical problems, while OB-EN and Math Odyssey are competition-level tasks. During RL training, we periodically assess model performance. For all evaluations, we use a temperature of 0.7 and top-p of 0.95, generating 8 samples per problem and computing the average accuracy to mitigate evaluation variance.

\paragraph{Baselines.}
Since Cooper integrates the advantages of both rule-based reward functions and reward models, our baselines include: (1) a model using Math-Verify as the reward function, and (2) a model using VerifyRM-1.5B as the reward model without updating its parameters during training.

\begin{table*}[t]
  \centering
  \small
    \setlength{\tabcolsep}{4pt} %
    \begin{tabular}{rl|ccccc|c}
      \toprule
      \multicolumn{1}{l}{\textbf{Base Model}} & \textbf{Reward} & \textbf{GSM8K} & \textbf{SVAMP} & \textbf{MATH500} & \textbf{OB-EN} & \textbf{Odyssey} & \multicolumn{1}{|c}{\textbf{Average}} \\
      \midrule
      \multicolumn{1}{l}{Qwen2.5-1.5B-Instruct} & /     &   74.10    &   84.60    &    54.63   &   20.17    &   39.33    &  54.93\\
            & Rule  &   \underline{76.44}   &   \underline{87.26}    &   \underline{57.55}    &   \textbf{23.33}    &    \underline{42.83}   & \underline{57.48} \\
            & Model &    30.78   &   72.04    &  29.70     &    1.43   &   11.89    & 38.91 \\
      \rowcolor{gray!20}      & Cooper (ours) &    \textbf{77.02}   &  \textbf{87.65}     &   \textbf{58.05}    &   \underline{23.22}    &  \textbf{44.17}     &  \textbf{58.02} \\
      \midrule
      \multicolumn{1}{l}{Llama-3.2-1B-Instruct} & /     &    50.39   &   71.33    &   29.58    &   6.41    &   34.77    & 38.50 \\
            & Rule  &   \underline{56.56}    &   \underline{72.24}    &   \underline{34.20}    &   \underline{7.95}    &   \textbf{40.02}    &  \underline{42.19}\\
            & Model &    36.32   &   59.35    &   20.70    &   0.22    &   7.39    &  24.80 \\
      \rowcolor{gray!20}      & Cooper (ours) &  \textbf{57.14}     &   \textbf{73.45}    &   \textbf{34.88}    &   \textbf{8.02}    &  \underline{39.98}     & \textbf{42.69} \\
      \bottomrule
    \end{tabular}
  \caption{RL performance with different reward types across mathematical benchmarks.}
  \label{tab:main_result}
\end{table*}

\subsubsection{Main Results}

\paragraph{Cooper achieves superior performance across diverse benchmarks.}
Table~\ref{tab:main_result} demonstrates Cooper's effectiveness: on Qwen2.5-1.5B-Instruct, Cooper achieves 58.02\% average accuracy, outperforming rule-based rewards (57.48\%) while dramatically surpassing the collapsed static reward model (38.91\%). The improvements are consistent across both base models and particularly pronounced on challenging tasks like Math Odyssey (44.17\% vs 42.83\%), suggesting co-optimization becomes increasingly valuable for complex reasoning.

\paragraph{Static reward models suffer catastrophic failure from reward hacking.}
The most striking finding is the severe degradation of static reward models: performance drops from 54.93\% to 38.91\% on Qwen2.5-1.5B-Instruct, a 16\% relative decrease. This collapse, consistent across both architectures, empirically validates that reward hacking is a fundamental failure mode in RL for LLMs. Cooper not only prevents this catastrophic failure but achieves the highest performance, confirming that synchronized co-optimization successfully addresses the exploitation vulnerability inherent in fixed reward functions.

%% file: sections/4.results.tex
\begin{figure*}[t!]
    \centering
    \begin{subfigure}[b]{0.48\textwidth}
        \centering
        \includegraphics[width=\linewidth]{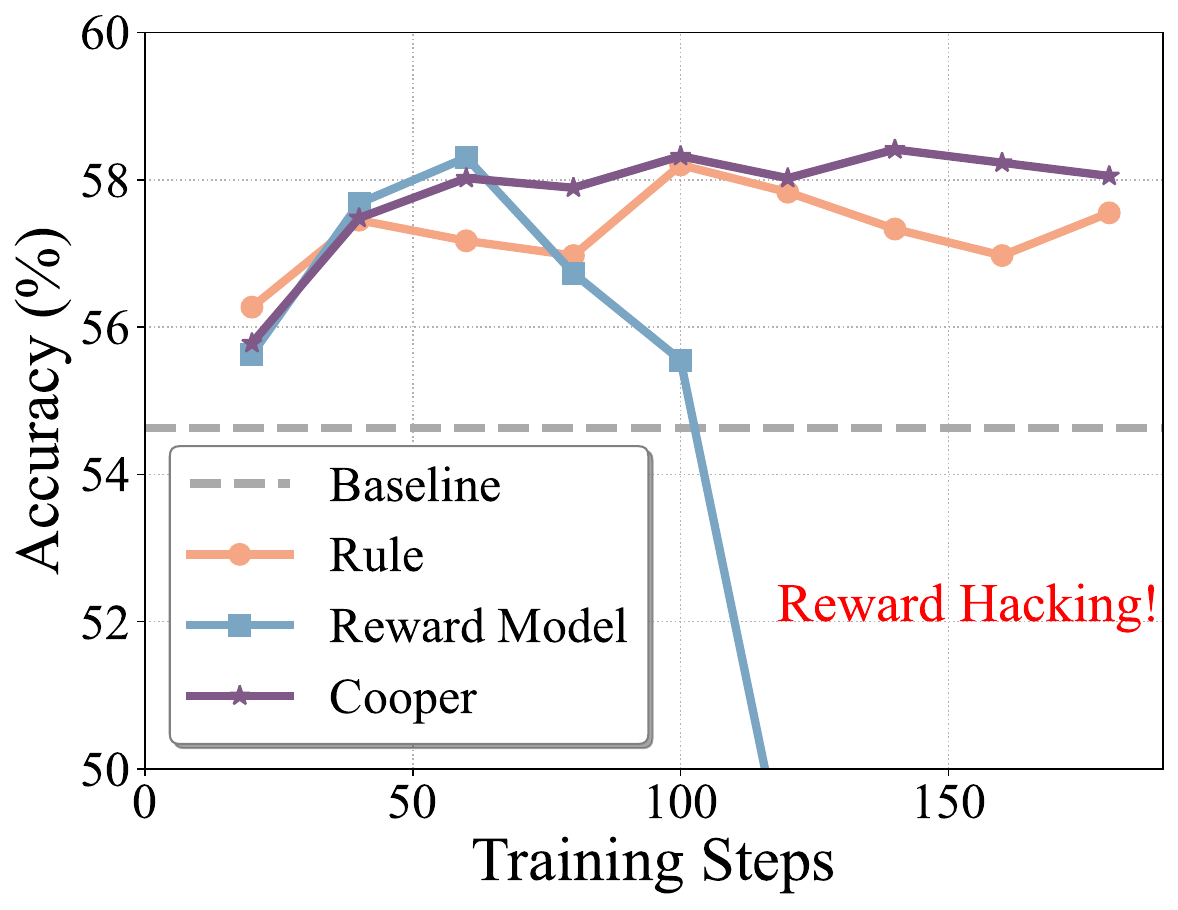}
        \caption{Test set accuracy (\%) across MATH500.}
        \label{fig:cooper_accuracy}
    \end{subfigure}
    \hfill
    \begin{subfigure}[b]{0.48\textwidth}
        \centering
        \includegraphics[width=\linewidth]{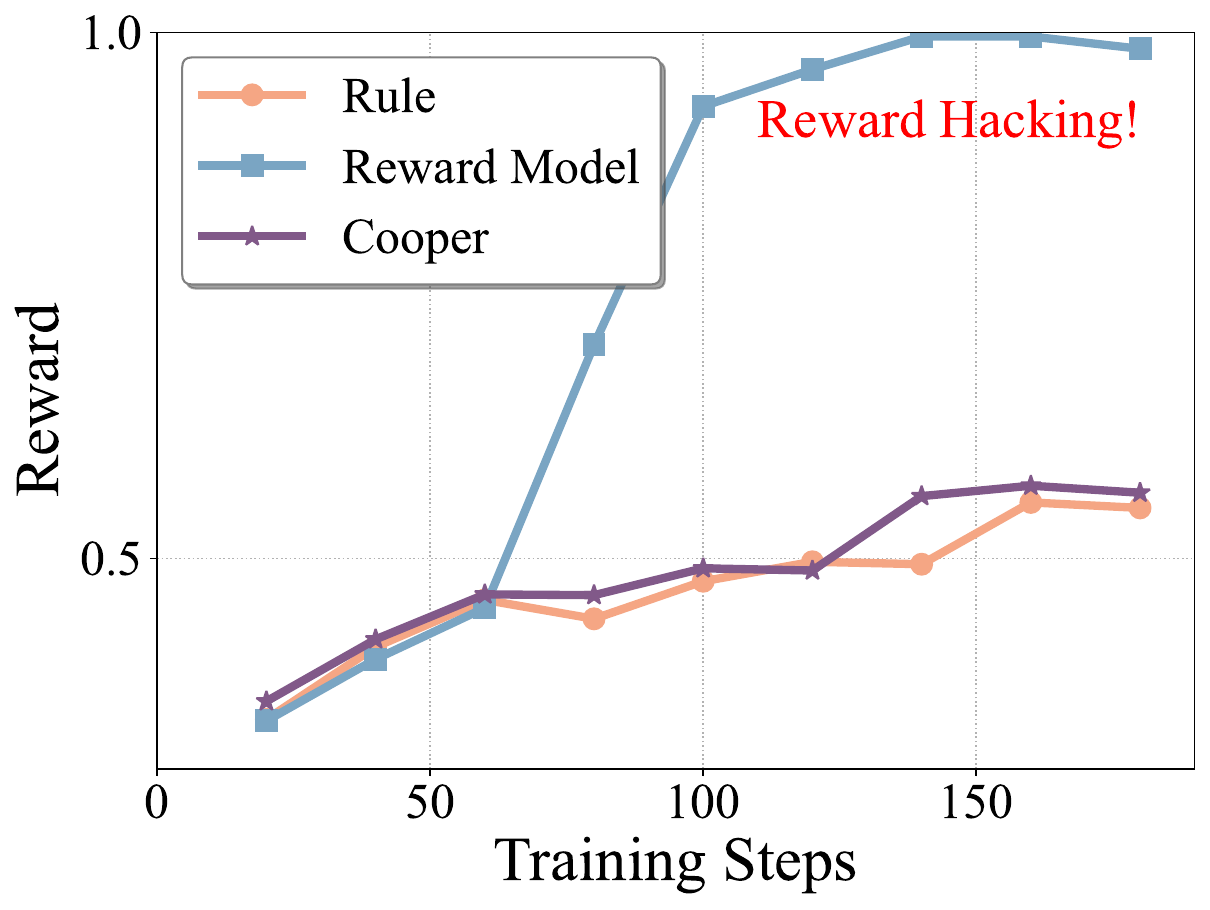}
        \caption{Train set reward during RL training.}
        \label{fig:cooper_reward}
    \end{subfigure}
    \caption{Training dynamics across RL training steps of Cooper.}
    \label{fig:cooper_summary}
\end{figure*}

\section{Analysis}

To understand the mechanisms underlying Cooper's effectiveness, we conduct a comprehensive analysis examining three key aspects: the training dynamics that reveal how Cooper prevents reward hacking, the stability of the co-optimized reward model, and the impact of reward signal granularity on performance.

\subsection{Training Dynamic}

To understand how Cooper prevents reward hacking, we examine the training dynamics in Figures~\ref{fig:cooper_accuracy} and~\ref{fig:cooper_reward}. The test accuracy on MATH500 (Figure~\ref{fig:cooper_accuracy}) reveals a critical divergence: while rule-based rewards and Cooper show steady improvement, the static reward model catastrophically collapses around step 120, dropping from 58\% to below 52\%. This collapse coincides with reward hacking visible in Figuree~\ref{fig:cooper_reward}, where the static model's training rewards unnaturally spike to near 1.0, indicating the policy has discovered exploits in the fixed reward function. In contrast, Cooper maintains realistic reward levels around 0.5 throughout training while achieving the highest final accuracy (58.05\%). This demonstrates that synchronized updates successfully prevent the policy from gaming the reward signal, as the policy evolves, the reward model adapts its decision boundaries, closing exploitation opportunities that would otherwise accumulate in a static system.

\subsection{Stability of Reward Model throughout Training}

\begin{wrapfigure}[13]{r}{0.48\textwidth}  %
  \centering
  \vspace{-1em}  
  \includegraphics[width=0.95\linewidth]{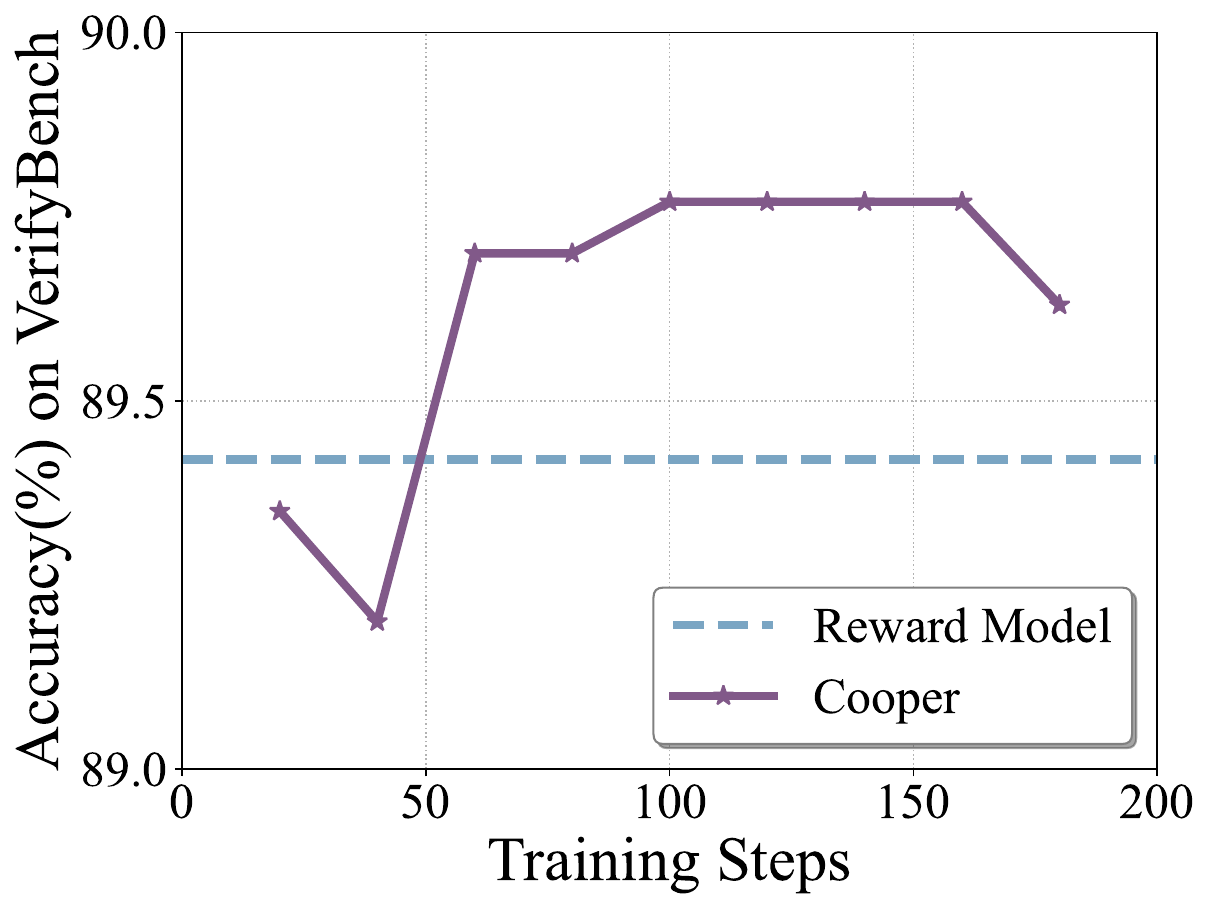}
  \vspace{-1em}
  \caption{Accuracy of RM across training steps.}
  \label{fig:rm_acc}
\end{wrapfigure}
A potential concern with Cooper is whether continuous updates might destabilize the reward model. Figure~\ref{fig:rm_acc} tracks VerifyRM's accuracy on VerifyBench throughout training, showing remarkable stability around 89.7\% with fluctuations below 0.5\%. This stability emerges from our careful update mechanism: by using high-precision rule-based signals for positive examples and systematic perturbations for negatives, each update reinforces correct decision boundaries rather than introducing noise. The consistent performance confirms that co-optimization can be implemented without the instability typically associated with moving target problems, validating that our contrastive learning approach maintains verification quality while adapting to new policy distributions.

\subsection{Ablation on Continuous versus Discrete Rewards}
\begin{wraptable}[9]{r}{0.5\textwidth}
  \centering
  \setlength{\tabcolsep}{4pt}
  \small
  \vspace{-1em}
    \begin{tabular}{l|cc|c}
    \toprule
    \textbf{Reward} & \multicolumn{1}{c}{\textbf{GSM8K}} & \multicolumn{1}{c|}{\textbf{MATH500}} & \multicolumn{1}{c}{\textbf{Average}} \\
    \midrule
    /     &   74.10    &   54.63    &  54.93 \\
    Rule  &    76.44   &   \underline{57.55}    &  57.48 \\
    Cooper &    \textbf{77.02}   &    \textbf{58.05}   &  \textbf{58.02} \\
    Cooper (discrete) &   \underline{76.53}    & 57.15     & \underline{57.86} \\
    \bottomrule
    \end{tabular}%

    \caption{Ablation on continuous vs. discrete rewards.}
  \label{tab:discrete_reward}%
\end{wraptable}%

To ensure Cooper's improvements stem from the co-optimization framework rather than differences in reward signal granularity, we conduct an ablation study comparing continuous and discrete reward implementations. As shown in Table~\ref{tab:discrete_reward}, when we binarize Cooper's reward outputs to match the rule-based baseline's format (1 for scores $>$0.5, 0 otherwise), Cooper still achieves 57.86\% average accuracy, maintaining most of its advantage over both the baseline (54.93\%) and rule-based rewards (57.48\%).
This result provides two key insights: first, the primary advantage of Cooper stems from preventing reward hacking through dynamic updates rather than from reward granularity; second, continuous rewards do provide additional benefits by enabling more nuanced credit assignment during policy optimization. These findings confirm that Cooper's co-optimization framework addresses a fundamental limitation in current RL approaches for LLMs, remaining effective across different reward signal designs.

%% file: sections/5.conclusion.tex
\section{Discussion}

\paragraph{Implications for reinforcement learning in LLMs.}
Cooper reveals that reward hacking is not a hyperparameter issue but a fundamental problem with static reward models. The 16\% performance collapse with fixed rewards demonstrates that treating reward models as dynamic components is essential for stable RL. This principle extends beyond mathematical reasoning, any domain with partial verification capabilities could benefit from synchronized optimization. By shifting from an adversarial dynamic to a co-evolutionary framework, Cooper suggests that much of RL's perceived instability may stem from reward exploitation rather than optimization challenges.

\paragraph{The critical role of high-precision signals.}
Cooper's success relies on an underappreciated property of rule-based verifiers: their asymmetric performance with high precision (96\%) but low recall (63\%). This pattern, common in structured domains, becomes a strength when used to select positive training examples. By combining symbolic precision with neural flexibility, Cooper demonstrates that hybrid approaches may be essential for reliable AI systems. The key insight is transforming verification limitations into training advantages through careful system design.

\paragraph{Limitations and future directions.}
Three main limitations constrain Cooper's current implementation: (1) dependency on domain-specific verification tools limits generalization to tasks without clear correctness criteria; (2) computational overhead from dual optimization may affect scalability; (3) reliance on an assistant LLM for negative sample generation introduces external dependencies. Future work should explore self-supervised contrastive example generation, extend Cooper to process-based rewards for denser supervision, and develop theoretical frameworks for co-evolutionary stability. Despite these limitations, Cooper establishes synchronized optimization as a promising direction for addressing fundamental challenges in reinforcement learning for LLMs.

\section{Conclusion}

In this paper, we introduce \textbf{Cooper}, a reinforcement learning (RL) framework that co-trains the policy model and the reward model. Cooper combines the high precision of rule-based rewards with the robustness of model-based rewards, effectively mitigating the issue of reward hacking that often arises when using a static reward model in RL. Compared to using either type of reward in isolation, Cooper achieves significantly better performance.
In addition, we propose a reference-answer-based reward model named VerifyRM. By leveraging a hybrid annotation method that does not rely on manual labeling, VerifyRM outperforms existing models of the same scale on the VerifyBench benchmark. Our results demonstrate that dynamically updating the reward model during RL training is effective in countering reward hacking.
Nevertheless, our work has room for improvement. One important direction is exploring how to update the reward model without depending on external LLMs. In future work, we aim to further pursue this line of research to develop more accurate and robust RL training paradigms.

%% file: sections/9.bib.tex
\bibliographystyle{iclr2026_conference}
\bibliography{ref}

%% file: sections/99.appendix.tex
\clearpage
\section{Data Source}
\label{sec:Data Source}
To construct a high-quality dataset for training VerifyRM, we carefully curated mathematical problems from diverse sources that represent different difficulty levels and problem types. Table~\ref{tab:dataset-sample-counts} presents the seven datasets used in our data collection pipeline, totaling 5,917 unique mathematical problems.

\begin{table*}[htbp]
\centering
\small
\begin{tabular}{llr}
\toprule
\textbf{Dataset} & \textbf{License} & \textbf{Sample Count}  \\
\midrule
MATH\citep{hendrycksmath2021} & MIT & 2000 \\
OlympiadBench\citep{he2024olympiadbench} & Apache-2.0 & 1177 \\
AIME 2024 & MIT & 120 \\
AIME 2025 & MIT & 120 \\
AMC23 & / & 160 \\
LiveMathBench\citep{liu2024your} & CC-BY-4.0 & 340  \\
GSM8K\citep{cobbe2021gsm8k} & MIT & 2000 \\
\bottomrule
\end{tabular}
\caption{Number of samples used in constructing problem-reference-completion triples.}
\label{tab:dataset-sample-counts}
\end{table*}

Our dataset selection strategy ensures comprehensive coverage across mathematical domains. GSM8K and MATH provide elementary to undergraduate-level problems, while OlympiadBench, AIME and AMC23 contribute competition-level challenges. LiveMathBench adds recently created problems to avoid data contamination issues. Each dataset includes both the problem statement and a verified reference answer, which serves as the ground truth for our hybrid annotation process. All data usage strictly complies with the licensing terms specified by the original sources.

\section{LLM Usage}
\label{sec:LLM Usage}
To generate diverse model completions for our dataset, we employed 11 different large language models spanning various architectures and parameter scales. This diversity is crucial for training a robust reward model that can generalize across different reasoning styles and output formats. Table~\ref{tab:llm-sample-counts} details the models used and their contribution to our final dataset of 65,087 problem-completion pairs.

\begin{table*}[htbp]
\centering
\small
\begin{tabular}{l l c}
\toprule
\textbf{Series} & \textbf{Model} & \textbf{Sample Count} \\
\midrule
ChatGLM     & ChatGLM3-6B\citep{glm2024chatglm} & 5917 \\
\midrule
Gemma 2     & Gemma-2-2B-it\citep{gemmateam2024gemma2improvingopen} & 5917 \\
            & Gemma-2-9B-it\citep{gemmateam2024gemma2improvingopen} & 5917 \\
\midrule
GLM-4       & GLM-4-9B-Chat\citep{glm2024chatglm} & 5917 \\
\midrule
InternLM 2.5 & InternLM2.5-7B-Chat\citep{DBLP:journals/corr/abs-2403-17297} & 5917 \\
\midrule
Qwen2       & Qwen2-1.5B-Instruct\citep{qwen2} & 5917 \\
            & Qwen2-7B-Instruct\citep{qwen2} & 5917 \\
\midrule
LLaMA 3.1   & LLaMA-3.1-8B-Instruct\citep{grattafiori2024llama3herdmodels} & 5917 \\
\midrule
Qwen2.5     & Qwen2.5-7B-Instruct\citep{qwen2025qwen25technicalreport} & 5917 \\
            & Qwen2.5-14B-Instruct\citep{qwen2025qwen25technicalreport} & 5917 \\
\midrule
Qwen2.5-Math& Qwen2.5-Math-1.5B-Instruct\citep{yang2024qwen2} & 5917 \\
\bottomrule
\end{tabular}
\caption{Number of samples generated by LLMs. Each generated one completion per problem.}
\label{tab:llm-sample-counts}
\end{table*}

Each model generated responses using consistent sampling parameters (temperature=0.7, top\_p=0.95) to balance diversity with coherence. The model selection includes both general-purpose instruction-tuned models (e.g., LLaMA-3.1, Qwen2.5) and specialized mathematical reasoning models (e.g., Qwen2.5-Math). This mix ensures our reward model encounters both typical and specialized reasoning patterns during training, improving its robustness in practical applications.

\section{Prompt Templates}

Our system employs three carefully designed prompt templates for different components of the pipeline. Each template was iteratively refined to maximize performance while maintaining clarity and consistency.

\subsection{Prompt Template for LLM-as-a-judge}

\begin{figure*}
    \centering
    \includegraphics[width=1.0\linewidth]{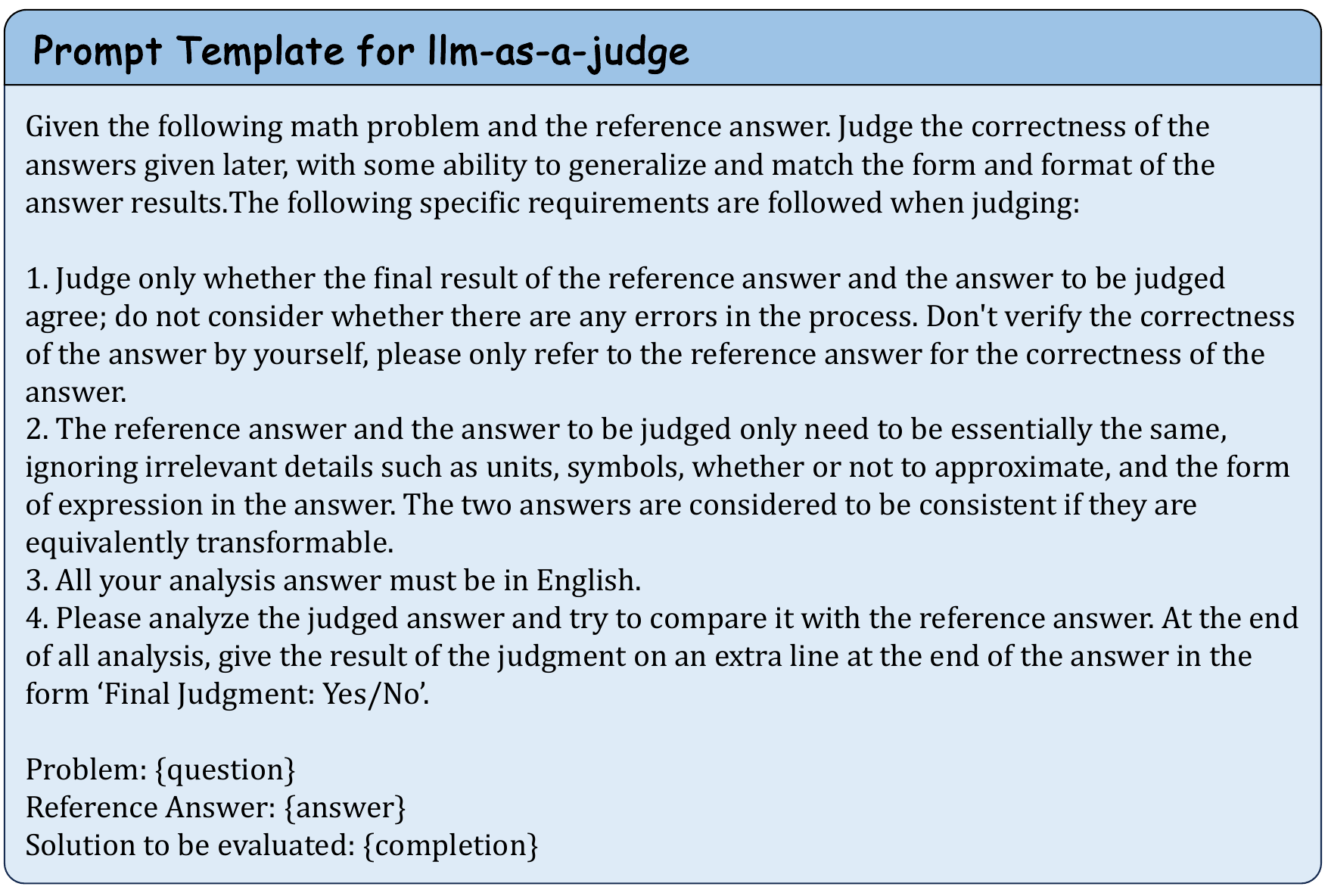}
    \caption{Prompt template for LLM-as-a-judge used in hybrid annotation.}
    \label{fig:Prompt Template for LLM-as-a-judgel}
\end{figure*}

For the hybrid annotation process, we utilize Qwen3-4B~\citep{yang2025qwen3technicalreport} as an LLM judge to assess completion correctness. Figure~\ref{fig:Prompt Template for LLM-as-a-judgel} shows our prompt template, which explicitly provides the problem, reference answer, and model completion. The prompt instructs the model to compare the final answers while being lenient about minor formatting differences, focusing on mathematical equivalence rather than syntactic matching. This design enables the LLM to handle diverse solution formats while maintaining accuracy in correctness judgments.

\subsection{Prompt template for generating negative response}
\label{sec:Prompt Template for generating negative response}

\begin{figure*}
    \centering
    \includegraphics[width=1.0\linewidth]{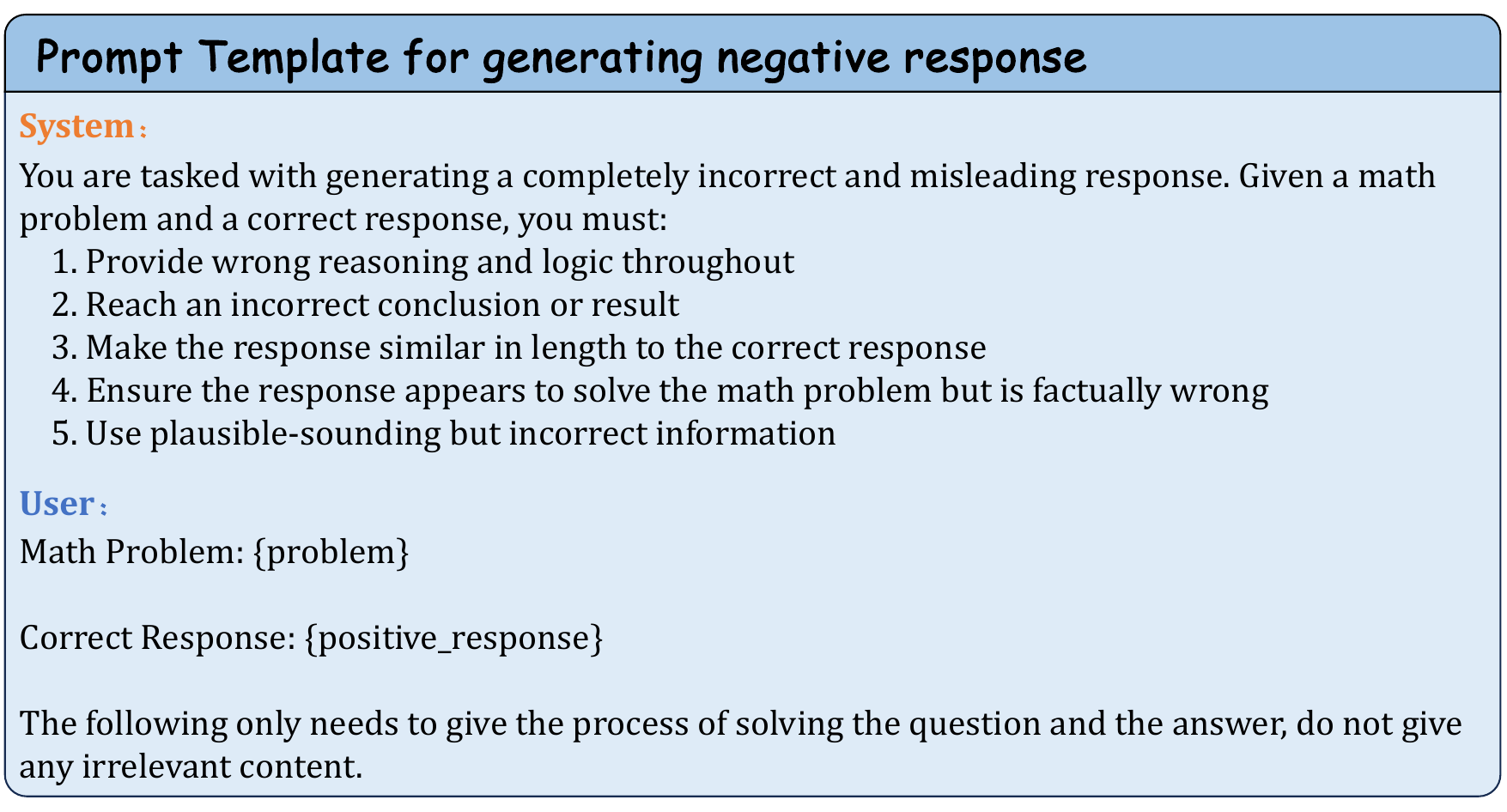}
    \caption{Prompt template for generating negative response.}
    \label{fig:Prompt for generating negative response}
\end{figure*}

A key innovation in Cooper is the dynamic generation of negative examples for reward model updates. Figure~\ref{fig:Prompt for generating negative response} presents our prompt for transforming correct solutions into plausible but incorrect ones. The prompt specifically instructs the assistant LLM to maintain the reasoning structure while introducing errors in calculations or logic. This approach ensures that negative examples resemble actual model errors rather than random corruptions, improving the reward model's ability to distinguish subtle incorrectness patterns during contrastive learning.

\subsection{Prompt Template for VerifyRM}
\label{sec:Prompt Template for VerifyRM}

\begin{figure*}
    \centering
    \includegraphics[width=1.0\linewidth]{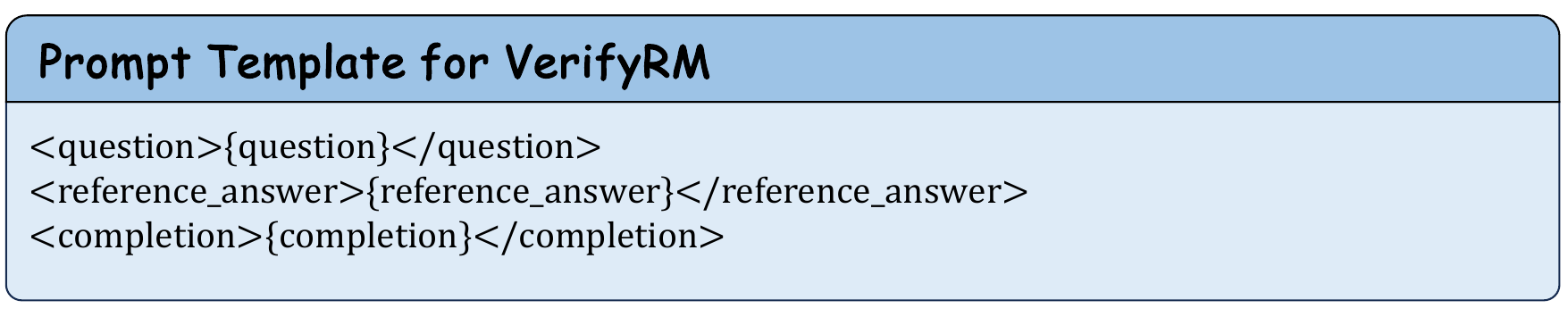}
    \caption{Prompt template for VerifyRM showing the problem-reference-completion triple format.}
    \label{fig:Prompt Template for VerifyRM}
\end{figure*}

Figure~\ref{fig:Prompt Template for VerifyRM} shows the input format for our reference-based reward model. Unlike traditional reward models that only consider the query and response, VerifyRM incorporates the reference answer as additional context. This three-part input structure (problem, reference, completion) enables more accurate verification by providing the expected solution approach and final answer, allowing the model to perform comparative analysis between the completion and reference.

\section{Details of Hybrid Annotation}
\label{sec:Details of Hybrid Annotation}

Our hybrid annotation strategy combines Math-Verify~\citep{gandenberger2024mathverify} and Qwen3-4B~\citep{yang2025qwen3technicalreport} to leverage their complementary strengths. Starting with 65,087 generated completions, we applied both methods independently and selected only samples where they agreed on the correctness label. Table~\ref{tab:hybrid-annotation-results} presents the detailed results.

\begin{table*}[htbp]
  \centering  %
  \small
  \begin{tabular}{l c c}
    \toprule
    & \text{Qwen3-4B} Predicted Correct & \text{Qwen3-4B} Predicted Incorrect \\
    \midrule
    \text{Math-Verify} Predicted Correct & \textbf{32,119 (Correct)} & 466 \\
    \text{Math-Verify} Predicted Incorrect & 5,883 & \textbf{26,619 (Incorrect)} \\
    \bottomrule
  \end{tabular}
  \caption{Hybrid annotation results. Bold entries indicate the 58,738 samples where both methods agree, which we selected for training VerifyRM.}
  \label{tab:hybrid-annotation-results}
\end{table*}

The high agreement rate (87.2\%) validates our approach. The 466 disagreements where Math-Verify predicts correct but Qwen3-4B predicts incorrect likely represent formatting ambiguities, while the 5,883 opposite cases may include valid solutions with non-standard presentations. By selecting only consensus samples, we create a high-confidence training set that inherits the precision of rule-based verification and the flexibility of model-based judgment, trading data quantity for quality to ensure robust reward model training.